\documentclass[11pt]{article}

\usepackage[margin=1in]{geometry}
\usepackage[T1]{fontenc}
\usepackage[utf8]{inputenc}
\usepackage{lmodern}
\usepackage{microtype}
\usepackage{amsmath,amssymb}
\usepackage{booktabs}
\usepackage{multirow}
\usepackage{graphicx}
\usepackage[numbers,sort&compress]{natbib}
\usepackage[colorlinks=true,linkcolor=blue,citecolor=blue,urlcolor=blue]{hyperref}
\usepackage{xcolor}
\hypersetup{
  pdftitle={How Much Velocity Does Off-Ball Space Value Need? A Broadcast-Viewport Benchmark},
  pdfauthor={Seongjin Choi},
  pdfsubject={Pitch control, player velocity, broadcast tracking, off-ball space value},
  pdfkeywords={football analytics, pitch control, velocity, broadcast tracking, imputation, benchmark}
}

\newcommand{\pp}{\,pp}          
\newcommand{\sv}{\sigma_v}
\newcommand{\zero}{\textsc{zero}}
\newcommand{\est}{\textsc{est}}
\newcommand{\vis}{\textsc{vis}}
\newcommand{\oracle}{\textsc{oracle}}

\title{How Much Velocity Does Off-Ball Space Value Need?\\A Broadcast-Viewport Benchmark}

\author{%
  Seongjin Choi\thanks{Independent researcher.
  ORCID: \url{https://orcid.org/0009-0001-3193-7424}.
  Code, logs and data: \url{https://github.com/nowayfootball/offscreen-impute}
  (directory \texttt{velocity/}); archived at \url{https://doi.org/10.5281/zenodo.22310851}.
  Contact: \texttt{nowayfootball@gmail.com}.}%
}
\date{}

\begin{document}
\maketitle

\begin{abstract}
Velocity-aware pitch control is standard, but under a
broadcast viewport half the players are off screen and on-screen velocities
come from a drifting calibration. We ask at which \emph{layer} of
broadcast off-ball analysis velocity changes the answer. Inheriting our
off-screen imputation protocol (three Metrica matches, 44\,m viewport,
block-bootstrap CIs), we score four velocity regimes --- none, viewport-legal
observed, true-for-visible, true-for-all --- against a velocity-aware
ground truth at three layers: imputation, the control surface, and
team verdicts. Velocity is nearly useless for imputation ($-0.2$\pp{}
against a 12--14\pp{} velocity-free surface MAE), first-order for the surface
($-1.5$ to $-1.8$\pp{}, 11--15\% of that MAE), and ten times
smaller for verdicts ($-0.12$ to $-0.19$\pp). The velocity that matters is the
visible channel: perfect occluded-player velocity adds 2--6\% of the visible
gain, and no last-seen decay policy we tested exceeds that. Omitting velocity blurs the surface (per-frame
$|e|$ 2.2--2.6\pp) with small time-averaged bias (per cell
$\le0.4$\pp), whereas imputation error is a structured bias
against the defending team's deep zone (5--9\pp). At a fixed velocity window,
a noise ladder of eleven jitter settings, including $\sv$-matched pairs, is
ordered to first order by one velocity-noise axis $\sv$ with break-even $\approx1$\,m/s; eleven SoccerNet-GSR clips from one
match through our pipeline measure $\sv=1.65$\,m/s yet recover 24--36\% of
the benefit: 43\% of the variance is frame-common, which the surface
tolerates, and the residual is heavy-tailed and clustered, which Gaussian
controls matched on component RMS do not reproduce ($+0.03$ vs.\ $+0.36$). The share of velocity-free error that velocity
removes grows with viewport width (7\% at 36\,m, 21\% at 60\,m): fix imputation on tight shots,
velocity on wide ones. Code and logs are released.
\end{abstract}

\section{Introduction}
\label{sec:intro}

Spatial football analytics on broadcast video has reached the point where a
single-camera feed can be converted into pitch coordinates for the players in
view~\citep{somers2024soccernet,gutierrez2024pnlcalib}, and the missing players
can be imputed well enough to compute pitch control and threat-weighted
territory~\citep{choi2026offscreen,choi2026junk}. Every such pipeline then
faces a design decision that the tracking-data literature never had to make
explicitly: \emph{how much effort should go into player velocity?} Physics-based
pitch control~\citep{spearman2017physics,spearman2018beyond} and
EPV~\citep{fernandez2021epv} take velocity as an input; on full-pitch optical
tracking it is essentially free. On broadcast video it is not. Off-screen
players have no observable velocity at all, on-screen velocities are finite
differences of positions that are themselves projected through a drifting
calibration, and commercial broadcast trackers report speed RMSEs ranging from
0.35 to 1.2\,m/s depending on the provider~\citep{crang2025broadcast}.

This paper does not propose a new pitch-control model or a new velocity
estimator. It measures, with a reproducible benchmark, \emph{where} velocity
changes the result of a broadcast-based off-ball analysis, and by how much.
We distinguish three layers at which a pipeline consumes velocity:

\begin{enumerate}
\item \textbf{Imputation} --- the positions of off-screen players (our earlier
  benchmark showed that blending last-seen velocity into the anchor improves
  position error by a fraction of a metre~\citep{choi2026offscreen});
\item \textbf{Surface} --- the per-cell pitch-control probability map;
\item \textbf{Verdict} --- scalar team-level aggregates of the surface, uniform
  (control share) or threat-weighted (expected-threat-weighted share, the
  quantity our possession-quality index actually reports~\citep{choi2026junk}).
\end{enumerate}

For each layer we ask how much is lost by setting velocity to zero, how much is
recovered by the velocity a broadcast pipeline can legally observe, and how
much headroom remains if occluded players were given their true velocity. We
then push the observed velocity through an observation-noise ladder, a
viewport-width sweep, and a real-broadcast axis on SoccerNet-GSR clips
processed by our own pipeline.

Our findings are:
\begin{itemize}
\item Velocity is a surface-layer input. It is nearly useless for imputation,
  first-order for the surface, and an order of magnitude smaller at the verdict
  layer (Section~\ref{sec:layers}). ``Cost'' throughout means distance from
  the velocity-aware reference surface of one Spearman-style model: the paper
  measures what velocity is needed to \emph{reproduce that reference}, not
  whether the reference is right about football.
\item The velocity that matters is already in the camera. Occluded-player
  velocity, even if perfect, adds 2--6\% of the visible-channel gain at the
  positions our imputers produce, and no decay constant we tested for
  last-seen velocity changes this (Section~\ref{sec:channels}).
\item The reason velocity does not reach the verdict is that omitting it
  \emph{blurs} far more than it \emph{biases}: per-cell time-averaged bias has
  RMS $\le0.14$\pp{} and never exceeds 0.4\pp{}, an order of magnitude below the per-frame error. Imputation error, by
  contrast, is a 5--9\pp{} structured bias against the defending team in its
  own deep zone (Section~\ref{sec:mechanism}).
\item At a fixed velocity window, eleven noise settings are ordered to first
  order by one velocity-noise axis, with break-even near 1\,m/s on three matches
  at 44\,m and on one match at 60\,m; real broadcast error sits above it in RMS but retains a quarter
  to a third of the benefit because its structure --- a frame-common component
  plus a heavy-tailed residual --- differs from the i.i.d.\ process (Sections~\ref{sec:noise}--\ref{sec:broadcast}).
\item The worth of velocity scales with how much of the pitch is visible
  (Section~\ref{sec:viewport}).
\end{itemize}

\section{Related Work}
\label{sec:related}

\paragraph{Velocity-aware space models.} \citet{spearman2017physics} and
\citet{spearman2018beyond} model pitch control from time-to-intercept with
reaction time and initial velocity; \citet{fernandez2021epv} build EPV on
velocity-aware influence surfaces. These are the models whose velocity term we
audit, not replace. At the other end, \citet{higgins2023pitchcontrol}
measure pitch control over six Premier League seasons with a velocity-free
Voronoi model; the surface-layer gap we report is a first estimate of what
that simplification costs under a viewport, measured by zeroing velocity in a
Spearman-style model rather than by evaluating their Voronoi implementation.

\paragraph{Velocity completion.} The closest prior task is that of
\citet{umemoto2025velocity}, who complete all 22 velocities from event
snapshots with no temporal information and show that space evaluation on
the completed velocities moves closer to the full-tracking result. Our setting is complementary: under a
broadcast viewport the visible players \emph{do} have observable velocity and
the occluded players do not, and our occluded-channel result (2--6\% of the
visible gain) is the marginal value of exact occluded velocities at the
positions our imputers produce. It is a marginal effect at fixed positions,
not a bound: Section~\ref{sec:channels} shows an undecayed stale velocity
beating the exact one on the hidden zone because it also compensates
positional lag, so a velocity predictor optimised for the surface at those
positions could exceed it.

\paragraph{Broadcast tracking accuracy.} \citet{crang2025broadcast} validate
three commercial broadcast trackers against TRACAB on a World Cup match and
report detection rates of 36--42\% on the programme feed, detected-player speed
RMSE of 0.35, 0.39--0.46 and 1.12--1.19\,m/s, and undetected-player speed RMSE
of 0.78--2.0\,m/s. These are the only published broadcast velocity-error
figures we know of and they are of the same order as our break-even, but they
cannot be placed on our axis: they are low-pass filtered, phase-aligned scalar
speeds, whereas our threshold is per-axis vector noise at a stated
finite-difference window. Ordering a provider against the threshold would
require its vector component errors under a matched filter.

\paragraph{Off-screen imputation.} This paper inherits the protocol,
imputation ladder and block-bootstrap of our off-screen imputation
benchmark~\citep{choi2026offscreen}, which scored imputers by position error and
by velocity-free pitch control; ghosting~\citep{le2017ghosting,yurko2024nflghosts}
and trajectory imputation~\citep{everett2023inferring,capellera2024transportmer}
are the learned alternatives whose velocity-related headroom we measure at
fixed positions.

\section{Benchmark Protocol}
\label{sec:protocol}

\paragraph{Data and viewport.} Three Metrica Sports sample matches
\citep{metrica2020}, first halves, 25\,Hz tracking sub-sampled to 5\,Hz
(13{,}490 evaluated frames per match). A virtual broadcast camera pans with an
exponential moving average of the ball's $x$ coordinate and reveals a 44\,m
window (36--60\,m in the viewport sweep); players outside it are hidden. This
is the protocol of \citet{choi2026offscreen}, unchanged.

\paragraph{Imputation.} Hidden players are imputed by the B2 team-centroid
relative-offset anchor (and B4, role-anchored voting) of the same benchmark;
we report B2 throughout as the imputer that isolates the velocity question most
cleanly, and note B4 where it differs.

\paragraph{Control model.} Pitch control follows \citet{spearman2017physics}:
each player's time to reach a cell is a reaction time $T_{\mathrm{react}}=0.7$\,s
during which they continue at their current velocity, plus distance at
$V_{\max}=7.8$\,m/s; team control is a logistic of the difference in minimum
arrival times with scale $0.45$\,s. The velocity term enters only through the
$T_{\mathrm{react}}\,v$ projection. Sensitivity to $T_{\mathrm{react}}$ and the
logistic scale is reported in Section~\ref{sec:sensitivity}.

\paragraph{Ground truth.} The reference surface uses all 22 true positions and
true velocities (5-frame differences of 25\,Hz tracking, capped at 9\,m/s).
Every regime below is scored against this same velocity-aware surface on the
same frames, so differences between regimes are paired.

\paragraph{Velocity regimes.} On the estimated side we run four regimes in one
process:
\begin{itemize}
\item \zero{}: all velocities set to zero (the velocity-free evaluation of
  \citet{choi2026offscreen});
\item \est{}: viewport-legal observed velocity --- finite differences of
  visible positions over one 0.2\,s step (over $k$ steps in the noise ladder);
  for hidden players the last-seen velocity decayed as
  $\exp(-\Delta t/\tau)$ with $\tau=1.5$\,s;
\item \vis{}: true velocity for visible players, zero for hidden;
\item \oracle{}: true velocity for all 22 (imputed positions, true velocities).
\end{itemize}
\paragraph{Cap policy.} Every velocity in every experiment --- reference,
visible observed, hidden last-seen, and both sides of the broadcast comparison
in Section~\ref{sec:broadcast} --- is capped at 9\,m/s. Noise-free, the cap
binds on 0.04\% of visible player-frames with a velocity estimate. In the
noise ladder $\sv$ is the nominal pre-cap value; the cap binds on 0.04\% of
visible player-frames at $\sv=0.7$\,m/s, 0.15\% at 1.23, 0.5\% at 1.84 and
1.3\% at 2.36, a negligible departure from nominal over the 0.6\,s-window
ladder. The exception is the 0.2\,s-window point
(nominal $\sv=4.24$\,m/s), where it binds on 14\%: its effective error is well
below nominal and its recovery of $-3.75$ understates the uncapped damage. The
broadcast RMS of 1.65\,m/s is measured after the cap.
\est{}$-$\zero{} is the cost of ignoring velocity; \vis{}$-$\zero{} the visible
channel; \oracle{}$-$\vis{} the occluded channel; and the \emph{recovery}
$(\est-\zero)/(\vis-\zero)$ the fraction of the visible-channel gain the
observed velocity delivers.

\paragraph{Metrics and layers.} Imputation layer: position error and the
B3V$-$B2 contrast (velocity-blended vs.\ pure anchor). Surface layer: control
MAE over the 3\,m grid, its hidden-zone restriction, and an xT-weighted MAE
using the same StatsBomb-trained expected-threat grid as
\citet{choi2026junk}. Verdict layer: absolute error of team control share and
of xT-weighted share (``spatial value'' error). Every paired contrast carries a
95\% one-minute block-bootstrap interval, and so does the recovery (the ratio
of means is recomputed in every bootstrap draw); other ratios of contrasts
(channel shares, fractions of \zero{} MAE) are ratios of point estimates and
are shown without intervals, and per-player position medians are descriptive. The intervals are conditional
on each match (blocks are resampled within a half); the three matches are
replications and are always reported separately, not pooled.

\paragraph{Noise ladder.} Only the estimated side's visible positions are
perturbed (ground truth and hidden-player scoring stay clean) with per-axis
Gaussian jitter of s.d.\ $\sigma\in\{0.3,0.6,1.0,1.5\}$\,m, either i.i.d.\ or
AR(1) with $\rho=0.9$ per 0.2\,s step (calibration-drift-like), and velocity is
taken over a $k$-step window ($k\in\{1,3,5\}$, i.e.\ 0.2/0.6/1.0\,s). The
induced per-axis velocity noise is
$\sv=\sigma\sqrt{2(1-\rho^k)}/(0.2k)$\,m/s. A common-mode arm splits the
jitter variance into a fraction $c$ shared by all visible players in a frame
and $1-c$ individual, preserving total variance.

\paragraph{Real-broadcast axis.} Eleven SoccerNet-GSR test clips
\citep{somers2024soccernet} (30\,s each, 25\,Hz pitch-coordinate ground truth
for visible players) are processed by our own broadcast pipeline
(detection, BoT-SORT tracking, PnLCalib-based
calibration~\citep{gutierrez2024pnlcalib}, stitching); detections
are Hungarian-matched to ground truth within 5\,m (match rate 67--97\%), and
only matched visible players with ground-truth team labels are scored, so the
axis isolates the position/velocity layer from team-assignment and detection
failures. This conditioning is deliberate and limiting: the axis says nothing
about the 3--33\% of players the pipeline misses. All eleven clips
(SNGS-116--126) are cut from a single match of the test split (game 7, both
halves, eleven separate moments), so the axis is one pipeline on one broadcast;
its intervals are a clip-level bootstrap (eleven clips resampled with
replacement, 2000 draws) and carry the variance across moments of that match,
not across matches, broadcasts or trackers.

\section{Results}

\subsection{Velocity is a surface-layer input}
\label{sec:layers}

Table~\ref{tab:layers} gives the paired \est{}$-$\zero{} contrast for the
three layers on B2.

\begin{table}[t]
\centering
\footnotesize
\caption{Cost of ignoring velocity by layer (B2 imputation, 44\,m viewport,
\est{}$-$\zero{}, percentage points, 95\% block-bootstrap CI). Imputation-layer
rows are the B3V$-$B2 contrast (velocity-blended vs.\ pure anchor) of
\citet{choi2026offscreen} re-scored against the velocity-aware reference in the
\zero{} regime; in the other three regimes the control-MAE contrast is
$-0.19$ to $-0.29$. Position error has no interval (it is a per-player
median).}
\label{tab:layers}
\setlength{\tabcolsep}{3.5pt}
\begin{tabular}{llccc}
\toprule
Layer & Metric & Match 1 & Match 2 & Match 3 \\
\midrule
Imputation & median position (m), B2$\to$B3V & 13.6$\to$13.2 & 12.8$\to$12.2 & 12.5$\to$11.9 \\
Imputation & control MAE, B3V$-$B2 & $-0.19$ [$-0.23,-0.15$] & $-0.20$ [$-0.25,-0.15$] & $-0.25$ [$-0.29,-0.20$] \\
Imputation & share error, B3V$-$B2 & $+0.07$ [$+0.03,+0.13$] & $+0.09$ [$+0.03,+0.13$] & $+0.01$ [$-0.04,+0.07$] \\
\midrule
Surface & control MAE & $-1.48$ [$-1.68,-1.26$] & $-1.68$ [$-1.79,-1.53$] & $-1.79$ [$-1.93,-1.63$] \\
Surface & hidden-zone MAE & $-0.33$ [$-0.40,-0.24$] & $-0.34$ [$-0.41,-0.27$] & $-0.42$ [$-0.51,-0.32$] \\
Surface & xT-weighted MAE & $-1.43$ [$-1.61,-1.21$] & $-1.61$ [$-1.72,-1.47$] & $-1.71$ [$-1.85,-1.56$] \\
\midrule
Verdict & share error & $-0.12$ [$-0.18,-0.05$] & $-0.19$ [$-0.27,-0.13$] & $-0.15$ [$-0.25,-0.06$] \\
Verdict & xT-weighted share err.\ & $-0.12$ [$-0.17,-0.06$] & $-0.19$ [$-0.26,-0.12$] & $-0.14$ [$-0.23,-0.05$] \\
\bottomrule
\end{tabular}
\end{table}

At the \textbf{imputation layer} velocity is a small term: blending last-seen
velocity into the anchor improves control MAE by 0.19--0.25\pp{} against a B2
control MAE of 12.1--13.7\pp{}, and median position error by 0.4--0.6\,m. It
does not improve the verdict --- the share error is slightly \emph{worse} under
B3V in matches 1--2 --- and the role-anchored B4 remains the best imputer by
position error under every velocity regime. For scale, the total verdict effect of imputation with
velocity held at truth for every player (\oracle{} regime, where true positions
give zero error by construction) is a control-share error of 6.1/5.4/4.4\pp{}
and an xT-weighted share error of 6.1/5.5/4.5\pp{} under B2; B4 brings the
former to 4.7 [3.6, 6.1] / 4.5 [3.6, 5.5] / 4.8 [3.8, 5.9], a paired
improvement of 0.7--2.1\pp{} in match 1 and 0.1--1.6 in match 2 (match 3
inconclusive). The imputer moves the verdict by an order of magnitude more
than any velocity term below.

At the \textbf{surface layer} velocity is a first-order term: ignoring it costs
1.5--1.8\pp{} of control MAE, 11--15\% of the B2 error and seven to eight and a
half times the imputation-layer term. Threat weighting does not change this (xT-weighted
MAE moves by the same amount).

At the \textbf{verdict layer} the effect is significant --- every interval
excludes zero --- but an order of magnitude smaller than at the surface,
$-0.12$ to $-0.19$\pp{} for both uniform and threat-weighted share. Our prior
expectation that threat weighting would re-expose the velocity effect (because
velocity errors concentrate near the ball, in high-xT zones) is rejected: the
xT-weighted share error moves by exactly as much as the uniform one.

\subsection{The velocity that matters is the visible channel}
\label{sec:channels}

\begin{table}[t]
\centering
\footnotesize
\caption{Channel decomposition of the surface-layer gain (B2 control MAE,
percentage points, 95\% CI). \vis{}$-$\zero{}: true velocity for visible
players only. \oracle{}$-$\vis{}: adding true velocity for occluded players.
Recovery: fraction of the visible-channel gain delivered by viewport-legal
observed velocity. The occluded/visible ratio is a ratio of means, no
interval.}
\label{tab:channels}
\begin{tabular}{lccc}
\toprule
 & Match 1 & Match 2 & Match 3 \\
\midrule
visible channel, \vis{}$-$\zero{} & $-1.49$ [$-1.69,-1.26$] & $-1.69$ [$-1.82,-1.55$] & $-1.81$ [$-1.95,-1.64$] \\
occluded channel, \oracle{}$-$\vis{} & $-0.06$ [$-0.13,+0.02$] & $-0.09$ [$-0.15,-0.04$] & $-0.04$ [$-0.10,+0.00$] \\
occluded / visible & 3.7\% & 5.2\% & 2.4\% \\
recovery $(\est-\zero)/(\vis-\zero)$ & 0.99 [0.99, 1.00] & 0.99 [0.98, 0.99] & 0.99 [0.99, 0.99] \\
\bottomrule
\end{tabular}
\end{table}

Table~\ref{tab:channels} splits the surface gain into channels. Giving visible
players their true velocity reproduces the whole \est{} gain; adding true
velocity for the occluded players on top moves control MAE by only
$-0.04$ to $-0.09$\pp{}, 2--5\% of the visible channel (B4: $-0.05$ to
$-0.11$, 3--6\%). Attaching a correct velocity to an incorrectly imputed position buys
almost nothing. The recovery of 0.99 is a consistency check rather than a
finding --- \est{}'s visible velocities are noise-free differences of true
positions --- and it is the ceiling for the noise ladder below.

\paragraph{Decay of last-seen velocity does not matter.} Sweeping the decay
constant of the hidden players' last-seen velocity, $\tau\in\{1.5, 5, \infty\}$,
plus $\tau=\infty$ with the 9\,m/s cap on \est{} velocities removed (reference velocities stay capped; match 1 only), moves \est{}$-$\zero{} from
$-1.48$ to at most $-1.58$; the largest occluded contribution any $\tau$ can
produce is $-0.08$\pp{}, 5.6\% of the visible channel, with a recovery
interval [0.94, 1.17] overlapping \vis{}. No decay policy takes the occluded
channel out of the 2--6\% bracket. The cleanest version of the same test is
$\tau=0$: observed velocity for visible players and \emph{zero} for hidden
ones, so that \est{} carries no hidden-velocity content at all. On all three
matches the observed point differences from the default \est{} are no larger
than 0.01\pp{}
(control MAE \est{}$-$\zero{} $-1.48/-1.68/-1.78$ against $-1.48/-1.68/-1.79$;
hidden-zone $-0.32/-0.34/-0.41$ against $-0.33/-0.34/-0.42$; recovery 0.99
on every match), and the direct paired contrast on the shared frames (default
minus $\tau=0$, block bootstrap) bounds the hidden last-seen contribution:
control MAE $-0.005$ [$-0.012,+0.007$], $-0.000$ [$-0.009,+0.009$] and
$-0.007$ [$-0.014,+0.002$]\pp{} on matches 1--3, hidden-zone MAE within
[$-0.022,+0.013$]\pp{}, verdict errors within [$-0.018,+0.007$]\pp{} (only
match 3's verdict intervals exclude zero, at $-0.008$ and $-0.006$\pp{}), and
recovery within [$-0.009,+0.005$] of the default.
The last-seen velocity of hidden players is worth no more than 0.02\pp{} on
any metric (largest interval bound 0.022), and the whole \est{} gain is the visible channel.
(With $\tau=\infty$ the undecayed velocity
happens to beat \oracle{} on the hidden zone, $-0.45$ vs.\ $-0.40$\pp{}, with
a wide interval; we read this as the $T_{\mathrm{react}}\,v$ projection partly
correcting the imputation's positional lag --- the same mechanism as B3V ---
rather than as velocity being valuable for occluded players per se.)

\subsection{Why velocity does not reach the verdict: blur far more than bias}
\label{sec:mechanism}

\begin{figure}[!htbp]
\centering
\includegraphics[width=\linewidth,height=0.45\textheight,keepaspectratio]{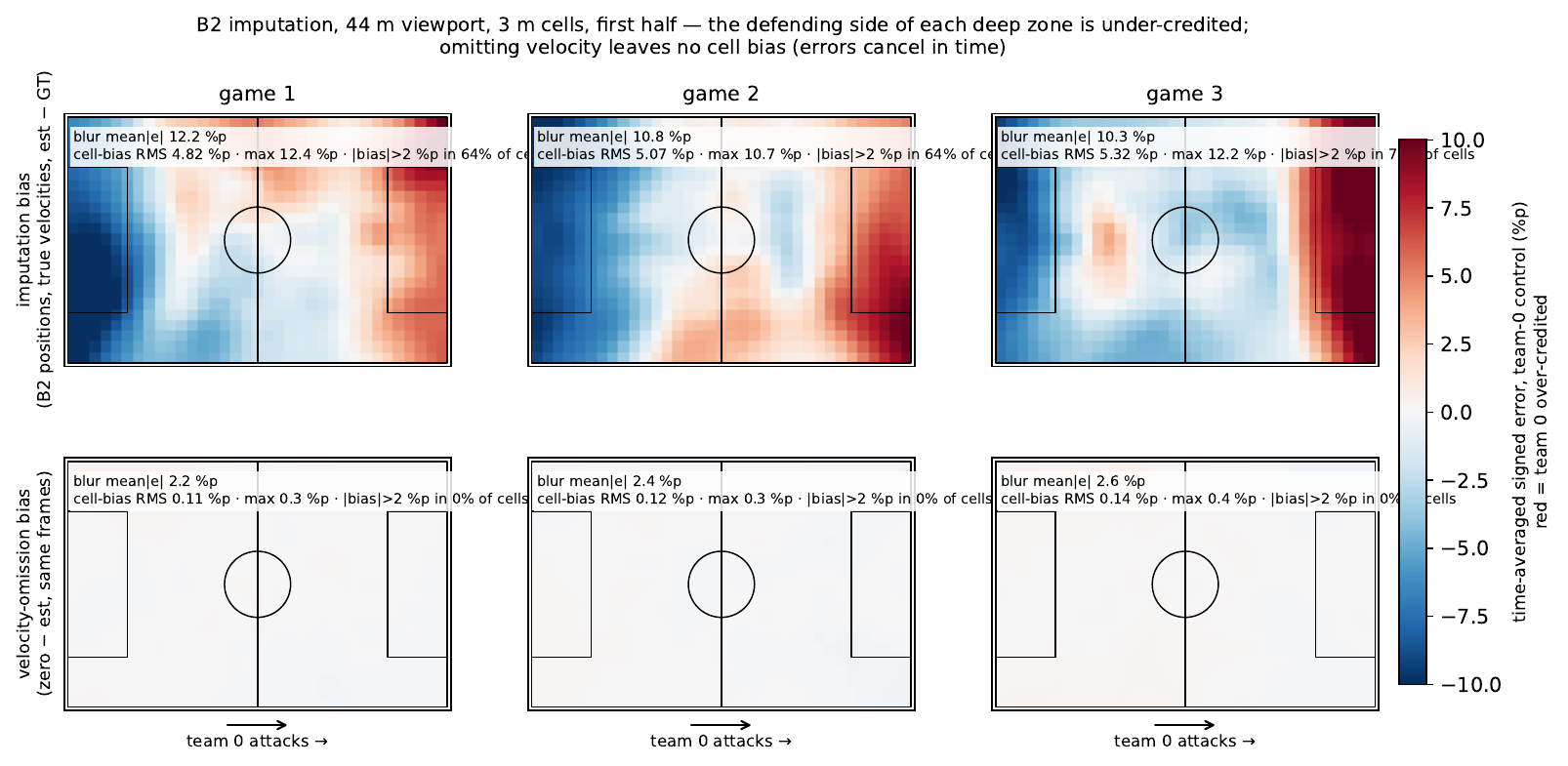}
\caption{Per-cell time-averaged signed control error (team 0, estimated minus
ground truth, 3\,m cells, first halves, B2), aligned so team 0 attacks right.
Top: imputation bias with true velocities for every player (\oracle{}
regime), i.e.\ the effect of B2 positions alone. Bottom: the velocity-omission
error, \zero{}$-$\est{}, on the same frames. Imputation error is a structured
bias that under-credits whichever team defends the deep zone; omitting
velocity leaves no cell with $|$bias$|$ above 0.4\pp{} despite a per-frame
$|e|$ of 2.2--2.6\pp{}.}
\label{fig:zonemap}
\end{figure}

Figure~\ref{fig:zonemap} and three per-frame statistics separate what
cancels where. The per-cell magnitude of the velocity-omission error is
2.25/2.43/2.63\pp{} per frame (matches 1--3). \emph{Within} a frame it partly
cancels spatially: the signed spatial mean of the same error --- the amount by
which omitting velocity shifts that frame's control share --- has mean
magnitude 0.80/0.86/0.84\pp{} under uniform weights and 0.81/0.86/0.84 under
xT weights, about a third of the per-cell magnitude, because a sprinting
player is under-credited ahead and over-credited behind. \emph{Over} time the
remainder cancels too: the per-cell time-averaged bias has RMS
0.11/0.12/0.14\pp{} with a largest cell of 0.28/0.34/0.39, and the
time-average of the per-frame share shift is $-0.02/-0.04/+0.04$\pp{}. What
reaches the verdict --- the time-average of the \emph{absolute} per-frame share
error --- is smaller still, $-0.12$ to $-0.19$\pp{} (Table~\ref{tab:layers}),
because the residual $\pm0.8$\pp{} per-frame shift $D$ rides on an
imputation share error $I$ of 4--6\pp{} whose sign it does not share: across
the three matches the two agree in sign in 47--52\% of frames and correlate
at $-0.03$ to $-0.04$ (B2; B4 similar). Decomposing the verdict cost
$\overline{|I+D|}-\overline{|I|}$ frame by frame, the 77--81\% of frames in
which $|I|>|D|$ contribute $\overline{\operatorname{sgn}(I)\,D}=-0.06$ to
$+0.02$\pp{} --- nothing net --- and the whole $+0.12$ to $+0.20$ (\zero{}
against \vis{}) comes from
the fifth of frames in which the velocity shift exceeds the imputation error.
The verdict's insensitivity to velocity is therefore two successive
reductions rather than additive shares: spatial cancellation takes the
per-cell 2.2--2.6\pp{} to a per-frame 0.8\pp{} (about two thirds), and
masking by a sign-independent imputation error takes 0.8 to 0.12--0.20; on a
surface with no imputation error, velocity omission alone would move a
frame's share by of order 0.8\pp{} (measured here at B2 positions). Neither
of the two weightings we tested re-exposes the time-averaged bias, and a
weighting built to concentrate on the worst cells could recover at most the
0.3--0.4\pp{} those cells carry. What survives is the per-frame absolute
error, which the xT-weighted MAE row of Table~\ref{tab:layers} measures
directly ($-1.4$ to $-1.7$\pp{}, the same as the uniform MAE). Omitting
velocity blurs the surface without tilting it.

Imputation error is of a different kind. Under every velocity regime
(\zero{}$\approx$\est{}$\approx$\vis{}$\approx$\oracle{}; velocity cannot
change this map; Fig.~\ref{fig:zonemap} shows the \oracle{} regime, the
\zero{} map differs by $<0.1$\pp{} in each deep zone) the per-cell bias has RMS
4.8/5.1/5.3\pp{}, 64--71\% of cells exceed 2\pp{}, and the structure is
identical in all three matches: the defending team is under-credited by
5--9\pp{} in the deep zone in front of its own goal (team 0's own end,
$x<18$\,m: $-8.4/-8.6/-7.0$; the far end, where team 1 defends:
$+5.4/+6.2/+9.1$ for team 0, i.e.\ team 1 under-credited by that amount). The anchor, pulled toward the visible teammates
near the ball, places the last line higher than it is and inflates the
attacking team's space in behind. That is the zone where xT is highest, so
\emph{threat-weighted verdicts are vulnerable to imputation, not to velocity}
--- and the target for learned occluded-player models is this bias, not blur.

\begin{figure}[!htbp]
\centering
\includegraphics[width=\linewidth,height=0.38\textheight,keepaspectratio]{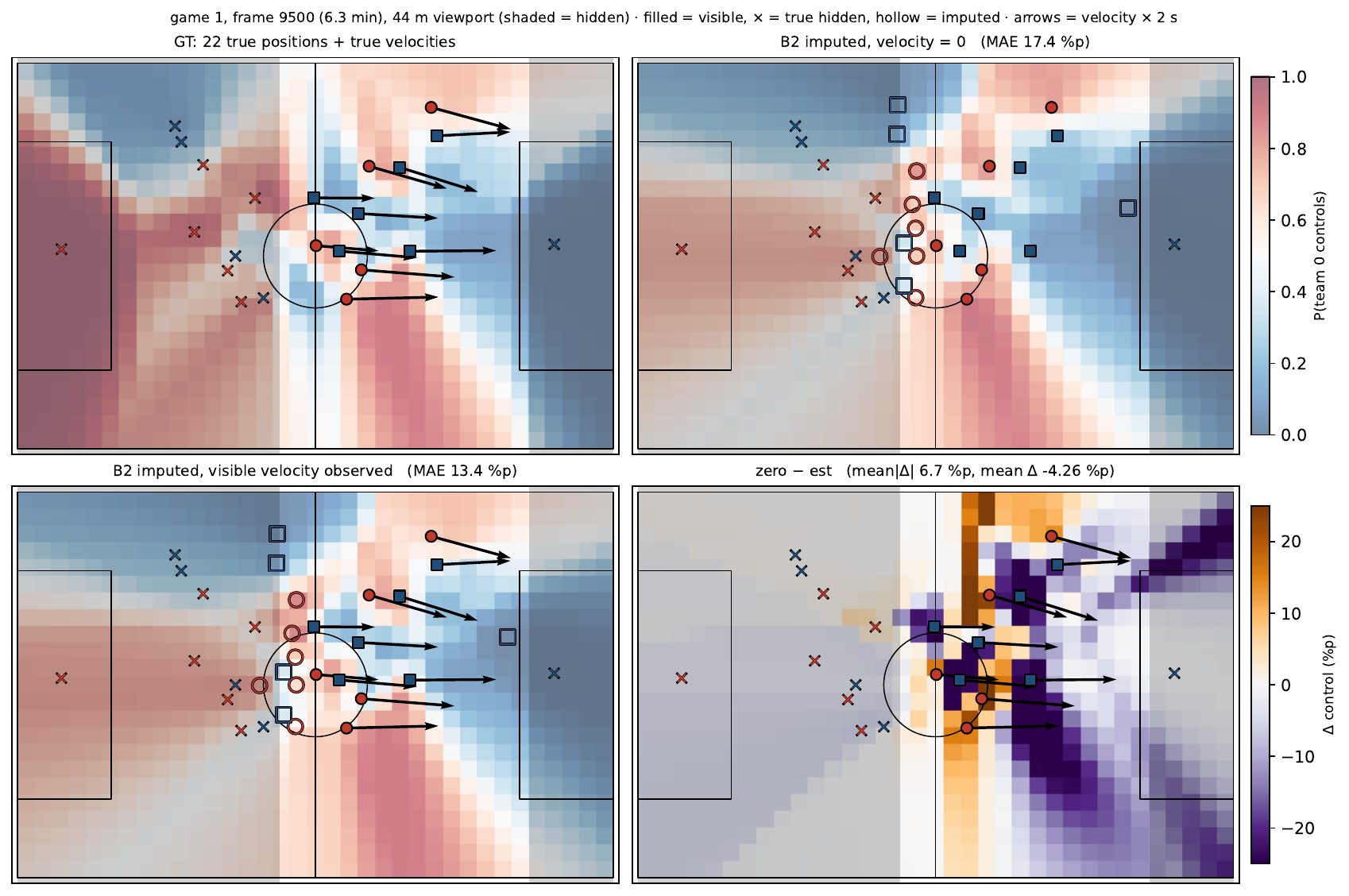}
\caption{One counter-attacking frame (match 1, 6.3\,min, 11 of 22 visible,
mean speed 5.6\,m/s over all 22 players). Top left: ground-truth surface. Top right: B2 with
velocity set to zero (MAE 17.4\pp). Bottom left: B2 with observed visible
velocity (13.4\pp). Bottom right: their difference. Within a frame the
velocity-omission error is one-sided (everything ahead of the sprint is
under-credited); over the half it averages to $\le0.4$\pp{} per cell. The
hollow markers show the imputation pulling the six hidden defenders toward the
viewport edge, emptying their own deep zone --- the bias of
Fig.~\ref{fig:zonemap}.}
\label{fig:case}
\end{figure}

Figure~\ref{fig:case} shows both mechanisms in a single frame.

\subsection{How much velocity noise the surface tolerates}
\label{sec:noise}

\begin{figure}[!htbp]
\centering
\includegraphics[width=\linewidth,height=0.40\textheight,keepaspectratio]{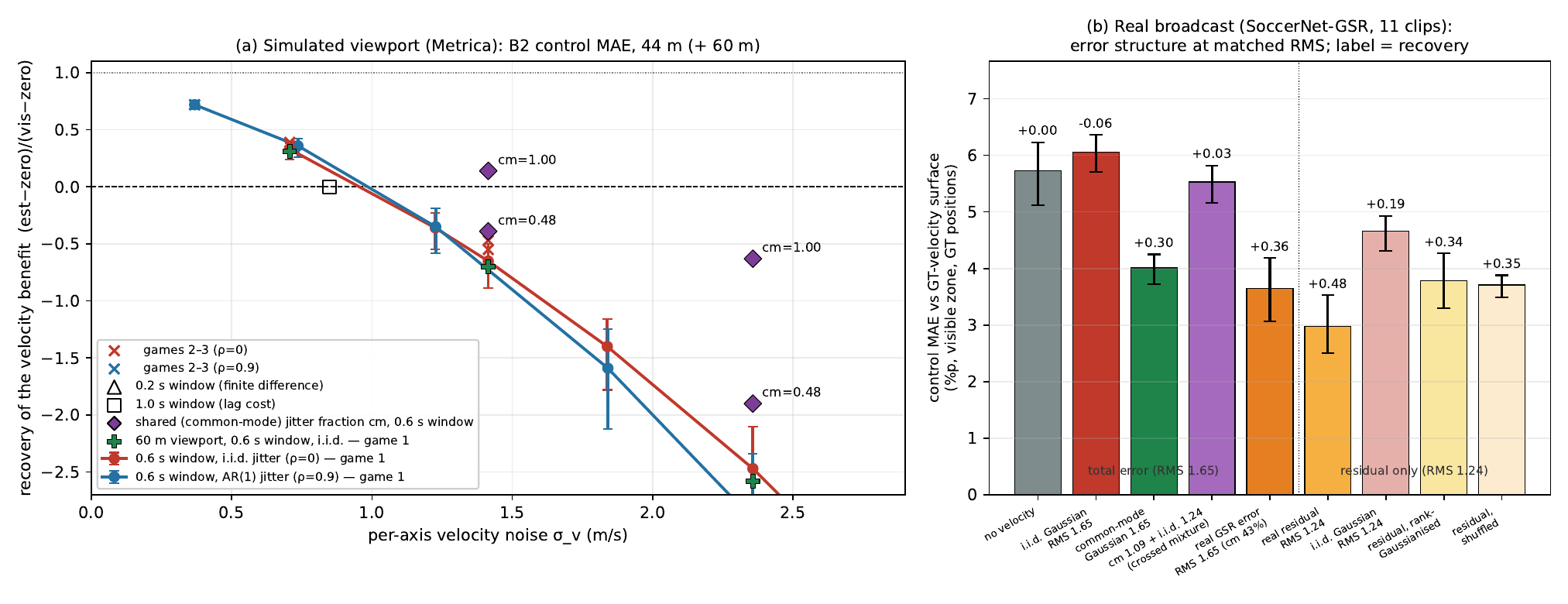}
\caption{(a) Recovery of the velocity benefit as a function of per-axis
velocity noise $\sv$ (match 1 lines, matches 2--3 crosses, 60\,m viewport
plus-markers). The eleven fixed-window settings ---
$\sigma\in\{0.3,0.6,1.0,1.5\}$\,m $\times$ $\rho\in\{0,0.9\}$ at 0.6\,s, plus
three settings chosen to match the $\sv$ of the other process
($\sigma=0.52$ and $0.78$\,m i.i.d., $\sigma=1.92$\,m at $\rho=0.9$) ---
are ordered to first order by $\sv$, with break-even at $\sv\approx 1$\,m/s
(paired differences of the matched pairs are given in the text); the two
other-window points (0.2\,s and 1.0\,s at $\sigma=0.6$\,m) are plotted on
the same axis but carry an additional window-dependent lag term; the 0.2\,s
point ($-3.75$) lies below the axis. Diamonds: common-mode arm at equal $\sv$. (b) The
real-broadcast axis: control MAE on 11 SoccerNet-GSR clips (clip bootstrap,
label = recovery). Left of the dotted line, errors at the total RMS 1.65: no
velocity, i.i.d.\ Gaussian, common-mode Gaussian, a crossed Gaussian mixture
with the measured component RMS values (common 1.09 + i.i.d.\ 1.24), and the
actual pipeline error. Right, the residual after removing the frame-common
component (RMS 1.24) against i.i.d.\ Gaussian at that RMS, the residual
rank-Gaussianised (tails removed, structure kept) and the residual shuffled
across samples (tails kept, structure removed); the re-centred $2\times2$
version of these controls is reported in the text.}
\label{fig:sigmav}
\end{figure}

Figure~\ref{fig:sigmav}(a) shows that, at a fixed 0.6\,s window, the recovery
of the velocity benefit is to first order a function of $\sv$ alone: i.i.d.\ and
AR(1) jitter at different $\sigma$ land on one curve, reproduced across the
three matches (recovery 0.72/0.72/0.72 at $\sv=0.37$; 0.33/0.36/0.39 at 0.71;
$-0.65/-0.55/-0.47$ at 1.41). The direct test is a pair of settings with the
same $\sv$ but different $\sigma$ and $\rho$, and the ladder contains four
(match 1): i.i.d.\ $\sigma=0.3$\,m ($\sv=0.71$) against AR(1) $\sigma=0.6$\,m
(0.74) recover 0.33 [0.24, 0.38] and 0.36 [0.26, 0.42]; i.i.d.\ 0.52\,m (1.23)
against AR(1) 1.0\,m (1.23), $-0.36$ [$-0.55,-0.23$] and $-0.35$
[$-0.58,-0.19$]; i.i.d.\ 0.78\,m (1.84) against AR(1) 1.5\,m (1.84), $-1.40$
[$-1.78,-1.16$] and $-1.59$ [$-2.12,-1.25$]; i.i.d.\ 1.0\,m (2.36) against
AR(1) 1.92\,m (2.36), $-2.47$ [$-3.05,-2.10$] and $-2.91$ [$-3.79,-2.34$]. In
every pair the position s.d.\ differs by a factor of two and the velocity
benefit itself (\vis{}$-$\zero{}) by about 0.25\pp. Because the members of a
pair share frames, the right test is the paired difference in recovery
(i.i.d.\ minus AR(1), block bootstrap on the shared frames): $-0.03$
[$-0.05,-0.02$] and $-0.01$ [$-0.06,+0.06$] at $\sv=0.71$ and 1.23, but
$+0.19$ [$+0.04,+0.38$] and $+0.45$ [$+0.16,+0.81$] at 1.84 and 2.36. Below
break-even the pairs are practically equivalent; above it the AR(1) member,
which carries the larger position variance, is measurably worse. So $\sv$
orders the ladder to first order, with a residual $\sigma^2$ dependence that is
negligible near and below break-even and material only where both members are
already far below zero. The mechanism is consistent: position noise enters all four
regimes identically --- they share the jittered positions --- so the recovery
isolates the velocity term at those positions; at a fixed window the
covariance between the current position error and the $k$-step velocity
error, $\sigma^2(1-\rho^k)/T=\sv^2T/2$, is itself fixed by $\sv$, and only the
position variance $\sigma^2$ is not.
Points from other windows do not lie exactly on this curve: the
1.0\,s-window point at $\sv=0.85$ recovers 0.00/0.04/0.09, about 0.1--0.2
below the fixed-window interpolation, of the same order as the lag cost
measured separately below; $\sv$ captures the jitter contribution only, not the
window's smoothing of true motion. It is also independent of the viewport: on a
60\,m window the same three noise levels give 0.31/$-0.70$/$-2.58$ against
0.33/$-0.65$/$-2.47$ at 44\,m (plus-markers in Fig.~\ref{fig:sigmav}a). A wider
window raises the velocity benefit and the noise cost in the same proportion
--- both scale with the number of visible players --- so the ratio does not
move. \textbf{Break-even is at $\sv\approx1$\,m/s} (fixed 0.6\,s window;
chord crossings 0.92--1.03) on the three matches at 44\,m and on match 1 at
60\,m, the only widths on which the ladder was run: above it, observed
velocity is worse than none. Three practical consequences follow.

\begin{itemize}
\item \emph{Window length is the first lever.} A 0.2\,s finite difference at
  $\sigma=0.6$\,m gives recovery $-3.75$ --- catastrophic, and with the 9\,m/s
  cap binding on 14\% of visible player-frames (Section~\ref{sec:protocol}) an
  understatement --- while a 0.6\,s
  window gives $-0.65$ and a 1.0\,s window $0.00$. But longer windows pay a lag
  cost even on clean data: at 1.0\,s the recovery is 0.75/0.73/0.72, i.e.\ the
  lag bias removes 25--28\% of the benefit. The window is a variance--lag
  trade-off whose optimum is a deployment parameter.
\item \emph{Temporal correlation is the second lever.} Calibration-drift-like
  AR(1) jitter at $\rho=0.9$ shifts $\sv$ down by a factor
  $\sqrt{1-\rho^k}$. Our own pipeline's matched position residuals have a
  lag-one (0.2\,s) autocorrelation of 0.84 (0.75--0.88 across clips,
  Section~\ref{sec:broadcast}), so for that pipeline the $\rho=0.9$ curve is
  the closer one. The position RMSE $\approx1.1$\,m alongside speed RMSE
  0.35\,m/s that \citet{crang2025broadcast} report is consistent with
  strongly correlated error, but, being a filtered scalar-speed metric, it
  cannot fix $\rho$.
\item \emph{The verdict is insensitive only conditionally.} Share error stays
  below the surface effect across the ladder, but its sign flips and becomes
  significantly worse at large $\sv$ (i.i.d.\ $\sigma=1.0$\,m:
  $+0.34$ [$+0.25,+0.44$]); ``verdicts are robust'' holds for
  $\sv\lesssim1$\,m/s.
\end{itemize}

The perfect-velocity gain itself shrinks with position noise
(\vis{}$-$\zero{} $-1.49\to-0.50$ at $\sigma=1.5$\,m): velocity is worth
having only when positions are right.

\paragraph{Common-mode error is neutralised, not converted.} Splitting the
jitter into a frame-common fraction $c$ at fixed $\sv$ (Fig.~\ref{fig:sigmav}a,
diamonds) raises recovery monotonically: at $\sv=1.41$ from $-0.65$ (i.i.d.)
to $-0.39$ ($c=0.48$, close to the 0.43 measured on the broadcast clips) to $+0.14$ [$+0.03,+0.23$]
($c=1$); at $\sv=2.36$ from $-2.47$ to $-1.90$ to $-0.63$. Even fully
common-mode jitter, on a full-pitch viewport with no hidden players, reaches
only $+0.04$ [$-0.06,+0.12$]: a shared velocity offset acts differently on
players moving toward and away from a cell, so relative arrival times do not
cancel it exactly. Common-mode error removes most of the damage; it does not
produce a gain. The hidden zone, which is the most fragile part of the surface
under i.i.d.\ jitter (recovery $-1.56$ and $-4.91$ at $\sv=1.41$ and $2.36$,
against $-0.65$ and $-2.47$ for the whole surface), is the most robust under
common-mode jitter ($+0.36$ and $-0.36$ at $c=1$, against $+0.14$ and
$-0.63$), even though imputed players do not receive the shared offset.

\subsection{The real-broadcast axis}
\label{sec:broadcast}

Our pipeline's matched visible-player error on the SoccerNet-GSR clips has
per-axis position s.d.\ 0.91\,m (0.61--1.16 across clips), lag-one (0.2\,s)
autocorrelation 0.84, and a $+0.2$ to $+0.65$\,m $y$ bias (a foot-point
convention difference that does not affect velocity). The resulting per-axis
vector velocity noise at a 0.6\,s window is $\sv=1.65$\,m/s, both as the RMS
pooled over all matched samples (1.649) and as the median of the eleven
per-clip RMS values (0.91--2.30). Measured against a true velocity formed with
the same trailing 0.6\,s window instead of the 0.4\,s central reference it is
1.47 (0.80--2.14), so about a tenth of the figure is the deterministic
window/phase discrepancy between the two operators rather than tracking
noise; both are well above the simulated break-even. Yet on the surface velocity
is still a gain: on true positions, pipeline velocity vs.\ zero gives
$-2.08$ [$-2.55,-1.56$]\pp{} (recovery 0.36); on pipeline positions,
$-1.60$ [$-2.14,-1.01$] (recovery 0.24); nine of eleven clips gain and two
(those with the largest position error, $\sv$ 1.85 and 2.30) break even. All of this is one pipeline on
eleven moments of one match; it is an existence result, not a survey of
broadcast trackers.

Figure~\ref{fig:sigmav}(b) resolves the apparent contradiction. I.i.d.\
Gaussian noise at the same RMS added to true velocities gives recovery
$-0.06$: no gain, the side of break-even the ladder predicts (the ladder's
magnitudes do not transfer, because here every scored player is visible at
its true position and the recovery denominator is the whole velocity-omission
error rather than its increment over imputation error). The real error
differs in structure. Subtracting
each frame's mean velocity error from every matched sample in that frame
splits the pooled per-sample variance exactly: 43\% is frame-common (RMS 1.09)
and 57\% is residual (RMS 1.24; $1.09^2+1.24^2\approx1.65^2$). We call the
first component frame-common rather than calibration drift: a shared
camera-parameter error would produce it, but so would coherent tracking
errors, and with a median of 12 matched players per frame an i.i.d.\ residual
would already place 5.6\% of the variance in frame means by chance (corrected
share 38\%). The residual is heavy-tailed, with median per-axis $|\Delta v|$
0.43\,m/s, 90th percentile 1.72, 99th 5.2, and 3.4\% of samples above
3\,m/s (median/RMS 0.35 against 0.67 for a Gaussian; the total error before
removing the common mode has median 0.53, 99th percentile 7.0 and 6.5\% above
3\,m/s). Common-mode Gaussian at 1.65 recovers $+0.30$; the real residual
alone recovers $+0.48$.

Controls matched on component RMS separate the two explanations. Common-mode
Gaussian at the measured 1.09 recovers $+0.52$ and the real frame-common
component alone $+0.63$ (a shared velocity offset translates the surface by
$T_{\mathrm{react}}\,v\le1$\,m and does little else); a crossed synthetic
mixture carrying both measured component RMS values, common-mode 1.09 plus
i.i.d.\ 1.24, recovers only $+0.03$, against $+0.36$ for the real error. The
residual's shape and structure carry the difference: i.i.d.\ Gaussian at the
residual's own RMS of 1.24 recovers $+0.19$ and the real residual $+0.48$. A
crossed $2\times2$ design separates the two: marginal distribution (empirical
vs.\ rank-Gaussianised) crossed with within-frame dependence (original
ordering vs.\ shuffled across all samples), with every cell re-centred per
frame and rescaled to RMS 1.24, so that the frame-common component and the RMS
are held fixed and only the residual's marginal and its dependence vary. The
cells recover $+0.48$ (empirical, original), $+0.35$ (Gaussian, original),
$+0.33$ (empirical, shuffled) and $+0.18$ (Gaussian, shuffled). In control-MAE
units with clip-bootstrap intervals, Gaussianising the marginal costs
$+0.83$\pp{} [0.72, 0.93], destroying the dependence $+0.91$\pp{} [0.51,
1.24], and the interaction is $+0.12$\pp{} [$-0.05$, 0.32]: the two factors are
close to additive ($-0.14$ and $-0.16$ in recovery units, $-0.30$ jointly) and
each accounts for about half of the gap between the real residual and an
i.i.d.\ Gaussian at its RMS. (The un-re-centred one-factor
controls in Fig.~\ref{fig:sigmav}b, $+0.34$ and $+0.35$, agree.) Most players at most moments
are tracked well, and the garbage is concentrated rather than spread.

The i.i.d.\ ladder is therefore conservative relative to the structured errors
we measured; we do not claim it is a worst case over all error structures. The
practical
guidance it yields is not ``measure $\sv$'' but ``measure the residual after
removing frame-common motion, and its tail, not only its RMS'': for this
pipeline a median $|\Delta v|$ of 0.43\,m/s delivered a surface gain at an RMS
of 1.65. Whether a robust-scale statistic predicts recovery across trackers
with different tail mass, bias and player correlation is untested here; the
controls above show that RMS alone under-predicts this pipeline's recovery by
a factor of 2.5.

\subsection{Velocity matters in proportion to what you can see}
\label{sec:viewport}

\begin{table}[t]
\centering
\small
\caption{Viewport-width sweep (match 1, B2 control MAE, percentage points,
95\% CI). The fraction of the velocity-free (\zero{}) error removed by true
visible velocity (ratio of means, no interval) rises with width; the
hidden-zone velocity gain and the recovery do not depend on it.}
\label{tab:viewport}
\footnotesize
\begin{tabular}{cccccc}
\toprule
Width & \zero{} MAE & \vis{}$-$\zero{} & gain / \zero{} & hidden-zone \vis{}$-$\zero{} & recovery \\
\midrule
36\,m & 16.86 & $-1.22$ [$-1.38,-1.03$] & 7\% & $-0.32$ [$-0.41,-0.22$] & 0.98 [0.96, 0.99] \\
44\,m & 13.73 & $-1.49$ [$-1.69,-1.26$] & 11\% & $-0.33$ [$-0.41,-0.25$] & 0.99 [0.99, 1.00] \\
52\,m & 11.26 & $-1.74$ [$-1.95,-1.49$] & 15\% & $-0.36$ [$-0.42,-0.27$] & 1.00 [0.99, 1.00] \\
60\,m & \phantom{0}9.04 & $-1.93$ [$-2.14,-1.68$] & 21\% & $-0.36$ [$-0.42,-0.30$] & 1.00 [1.00, 1.00] \\
\bottomrule
\end{tabular}
\end{table}

We expected narrower viewports to raise the share of the hidden-zone velocity
term. Table~\ref{tab:viewport} rejects this: the hidden-zone gain is flat
($-0.32$ to $-0.36$) and recovery is 0.98--1.00 at every width. What changes
is \emph{which channel is the bottleneck}. At 36\,m true visible velocity
removes 7\% of the velocity-free error; at 60\,m that error halves (16.9 to
9.0) and the fraction triples to 21\%. (The \zero{} MAE is not purely
imputation error --- it is positional error, velocity omission and their
interaction --- so we call it velocity-free error rather than imputation
error.) Investment in velocity
accuracy pays in proportion to how wide the shot is; on tight shots, fix
imputation first. The noise tolerance, by contrast, does not depend on width
(Section~\ref{sec:noise}): a wider shot makes velocity worth more, not easier
to get right.

\subsection{Sensitivity to the control model}
\label{sec:sensitivity}

\begin{table}[t]
\centering
\small
\caption{Sensitivity of the velocity benefit to the control model (match 1,
B2 control MAE, percentage points). Clean: no observation noise. Noisy:
$\sigma=0.6$\,m i.i.d., 0.6\,s window ($\sv=1.41$\,m/s). $T_{\mathrm{react}}$
rows are replicated on matches 2--3 in the text.}
\label{tab:sens}
\footnotesize
\begin{tabular}{lcccc}
\toprule
Setting & \vis{}$-$\zero{} (clean) & recovery (clean) & recovery ($\sv=1.41$) & \oracle{}$-$\vis{} \\
\midrule
$T_{\mathrm{react}}=0.4$\,s & $-0.87$ [$-0.98,-0.73$] & 0.99 [0.98, 1.00] & $-1.05$ [$-1.41,-0.84$] & $-0.02$ [$-0.07,+0.02$] \\
$T_{\mathrm{react}}=0.7$\,s (default) & $-1.49$ [$-1.69,-1.26$] & 0.99 [0.99, 1.00] & $-0.65$ [$-0.89,-0.50$] & $-0.06$ [$-0.13,+0.02$] \\
$T_{\mathrm{react}}=1.0$\,s & $-2.09$ [$-2.36,-1.77$] & 1.00 [0.99, 1.00] & $-0.50$ [$-0.70,-0.37$] & $-0.10$ [$-0.21,+0.00$] \\
\midrule
scale $=0.30$\,s & $-1.75$ [$-1.99,-1.47$] & 0.99 [0.98, 1.00] & $-0.64$ [$-0.88,-0.50$] & $-0.07$ [$-0.16,+0.03$] \\
scale $=0.45$\,s (default) & $-1.49$ [$-1.69,-1.26$] & 0.99 [0.99, 1.00] & $-0.65$ [$-0.89,-0.50$] & $-0.06$ [$-0.13,+0.02$] \\
scale $=0.60$\,s & $-1.29$ [$-1.45,-1.09$] & 0.99 [0.99, 1.00] & $-0.66$ [$-0.91,-0.50$] & $-0.04$ [$-0.11,+0.02$] \\
\bottomrule
\end{tabular}
\end{table}

Velocity enters the control model only through the $T_{\mathrm{react}}\,v$
projection, so the \emph{size} of the velocity benefit is close to linear in
$T_{\mathrm{react}}$ (Table~\ref{tab:sens}: $-0.87/-1.49/-2.09$ for
0.4/0.7/1.0\,s), and a sharper logistic raises it. None of this touches the
layer or channel conclusions: clean recovery is 0.99--1.00 and the occluded
channel stays within $-0.02$ to $-0.10$ in all six settings. What does move
is the break-even. Under the same $\sv=1.41$ the recovery is
$-1.05/-0.65/-0.50$ for $T_{\mathrm{react}}=0.4/0.7/1.0$: a longer reaction
window amplifies the benefit faster than the noise cost, so break-even rises
with $T_{\mathrm{react}}$. Matches 2 and 3 reproduce the ordering
($-0.94/-0.55/-0.41$ and $-0.81/-0.47/-0.34$), the clean gain
($T_{\mathrm{react}}=1.0$: $-2.37/-2.53$, 1.40$\times$ the default against a
nominal 1.43$\times$) and the occluded-channel contrast (2--6\%). At the one
noisy level tested ($\sv=1.41$, match 1) the logistic scale leaves recovery
unchanged ($-0.64/-0.65/-0.66$) --- it rescales benefit and cost together ---
though we did not re-estimate the whole curve or its zero crossing per scale. The
$\approx1$\,m/s break-even of Section~\ref{sec:noise} is therefore a statement
about $T_{\mathrm{react}}=0.7$\,s; a pipeline using a different reaction time
should read its own break-even off the $T_{\mathrm{react}}$ axis, on which the
scale parameter appears to matter little.

\section{Discussion: what this means for a deployed pipeline}
\label{sec:discussion}

\paragraph{Order of investment.} The results order the velocity-related
engineering of a broadcast pipeline. First, imputation: on a 36--44\,m shot
true visible velocity removes only 7--11\% of the velocity-free surface error,
and imputation is the only source of structured bias
(Section~\ref{sec:mechanism}). Second, the velocity \emph{window} and
calibration stability for visible players: the window is a variance--lag
trade-off with a 25--28\% lag cost at 1\,s and a catastrophe at 0.2\,s under
realistic jitter, and frame-common calibration drift is the benign component
of the error. Last, occluded-player velocity, which no decay policy and not
even an oracle lifts above 6\% of the visible channel.

\paragraph{Re-reading our earlier indices.} Our possession-quality index
\citep{choi2026junk} computes threat-weighted control from broadcast tracking
with 0.24\,s finite-difference velocities. On the i.i.d.\ ladder that setting
is far past break-even; the real-broadcast axis shows why the index was not
wrecked by it --- the error is mostly common-mode and heavy-tailed --- and the
verdict-layer result shows that the velocity error that did reach the surface
largely averaged out of the scalar index (Section~\ref{sec:noise} shows this
holds only below $\sv\approx1$\,m/s on the i.i.d.\ ladder; on the real axis
the index sits in the common-mode, heavy-tailed regime where it does). Our
reading --- that the index's velocity term was \emph{harmless} rather than
\emph{helpful} at that window, and that a 0.6\,s window would have made it
helpful --- is a hypothesis: we have not re-run the index at 0.24 and 0.6\,s on
its own data, and the real-broadcast axis here is at 0.6\,s. Conversely, the deep-zone imputation bias of
Section~\ref{sec:mechanism} is the term that index \emph{is} exposed to, and
it favours the attacking side of every deep zone.

\paragraph{What a learned occluded-player model should target.} For a model
that predicts occluded velocities on top of B2/B4 positions, the marginal
effect of getting them exactly right is the occluded channel: 2--6\% of the
visible gain. This is not a bound on a joint position--velocity predictor:
Section~\ref{sec:channels} shows an undecayed stale velocity beating the true
one on the hidden zone because it compensates positional lag, so a velocity
chosen to minimise surface loss at a wrong position need not be the true one.
The positional headroom is not --- it is a 5\pp{} RMS structured bias with a
consistent sign, and the imputer choice moves the verdict by 0.1--2\pp{} where
velocity moves it by 0.1--0.2. A learned imputer should be evaluated on the deep-zone bias
map, not on blur.

\section{Limitations}
\label{sec:limitations}

\begin{itemize}
\item \textbf{\est{} is an optimistic ceiling in simulation.} Its visible
  velocities are noise-free differences of true positions. The noise ladder and
  the real-broadcast axis are the correction, and the latter measures one
  pipeline (ours) on 5.5 minutes of footage.
\item \textbf{One provider, first halves, three matches.} As in
  \citet{choi2026offscreen}. Attack direction is estimated from the first
  50 frames. Statements of the form ``on every match'' or ``at every width''
  mean the three matches and four widths tested; the decay sweep is match 1
  only.
\item \textbf{One control model.} A Spearman-style arrival-time model with one
  parameterisation; Section~\ref{sec:sensitivity} shows the layer and channel
  conclusions are stable across $T_{\mathrm{react}}\in[0.4,1.0]$\,s and
  logistic scale $\in[0.3,0.6]$\,s, but the break-even $\sv$ is conditional on
  $T_{\mathrm{react}}$; $V_{\max}$ was not varied. Results are stated for this
  family, not for learned influence surfaces.
\item \textbf{Static xT grid.} Threat weighting ignores ball position and
  phase; ``value'' here is an xT approximation, not EPV.
\item \textbf{The interpretation of the imputation bias is a hypothesis.}
  That the anchor is compressed toward the visible teammates is consistent with
  the maps but not separately tested.
\item \textbf{Hidden-zone scoring exists only in simulation.} Real broadcast
  has no occluded-player ground truth; the real-broadcast axis scores visible
  players only, and its reference velocities are 0.4\,s central differences of
  annotated boxes with their own noise ($\approx0.5$\,m/s), so the measured
  $\sv$ includes reference noise and the 0.6\,s-trailing versus 0.4\,s-central
  operator mismatch (1.65 against 1.47 under a matched operator) and overstates
  the pipeline's own; the Gaussian control arms add noise to the reference
  velocity and so omit that window term; and the reference surface is itself
  imperfect. The clip-level bootstrap has eleven
  units from one match; its intervals are wide, their coverage is nominal at
  best, and they say nothing about variation across matches or trackers.
  Team-assignment errors (two clips with swapped teams in our pipeline) are a
  separate layer excluded here.
\end{itemize}

\section{Reproducibility}
\label{sec:repro}

All experiments are single commands on the released benchmark
(\texttt{impute\_bench.py}): \texttt{--vel paired} produces the four regimes
and paired intervals; \texttt{--zone-map} also writes the per-frame
share-shift diagnostic and \texttt{--frame-dump} the per-frame series behind
the sign decomposition; \texttt{--noise}, \texttt{--noise-rho},
\texttt{--vel-win} and \texttt{--noise-cm} the ladder, its matched pairs and
the common-mode arm;
\texttt{--hid-tau} and \texttt{--vcap} the decay sweep, with
\texttt{--hid-tau 0} giving the zero-hidden-velocity regime; \texttt{--width} the
viewport sweep; \texttt{--zone-map} and \texttt{--case-dump} the figures;
\texttt{--t-react}/\texttt{--sig} the sensitivity. The real-broadcast axis is
\texttt{p3\_e6\_broadcast.py}, \texttt{p3\_e6\_decompose.py} and
\texttt{p3\_e6\_structure.py} (matched-RMS controls) on the public
SoccerNet-GSR test clips; our pipeline's tracks for those clips are released
with the code, so every downstream number can be recomputed without the
pipeline. Logs for every number in this paper are in the repository's
\texttt{velocity/\_p3/} directory, \texttt{reproduce.sh} maps each log to
its command, and the figures are generated from those logs. The release is
archived at \url{https://doi.org/10.5281/zenodo.22310851}.

\section{Conclusion}
\label{sec:conclusion}

Reproducing a velocity-aware control surface from broadcast needs velocity at
the surface and an order of magnitude less of it at the verdict, and the
velocity it needs is already in the camera. Under a broadcast viewport,
occluded-player velocity is a 2--6\% term at the positions any of our imputers
produce,
velocity omission blurs with only a small time-averaged bias, and imputation error --- a
5--9\pp{} under-crediting of the defending team's deep zone --- is what threat-weighted verdicts are actually
exposed to. Observed velocity is worth using when its per-axis noise is below
about 1\,m/s in i.i.d.\ terms --- a threshold that was similar across the
three matches at 44\,m and between 44 and 60\,m on match 1 --- and our own
tracker on one broadcast exceeded that RMS bar (1.65\,m/s) yet still recovered
a quarter to a third of the benefit, because its error is a frame-common
component plus a heavy-tailed residual rather than the i.i.d.\ process the
bar was set on. For a broadcast pipeline the order of investment is
therefore: imputation first on tight shots, velocity window and calibration
stability second, occluded-player velocity last.

\setlength{\bibsep}{3pt}
\bibliographystyle{plainnat}
\bibliography{refs}

\end{document}